\documentclass{article}
\usepackage{spconf}
\usepackage{amsmath, amssymb, psfrag, bm, latexsym, color, amstext}
\usepackage{graphicx}
\usepackage{epstopdf, subfigure}
\usepackage{ upgreek }
\usepackage{ textcomp }
\usepackage{ dsfont }
\usepackage{booktabs}

\usepackage[utf8]{inputenc}

\title{Automatic Target Recognition Using \\ Discrimination Based on Optimal Transport}
\name{Ali Sadeghian$^{\star}$ \qquad Deoksu Lim$^{\star}$ \qquad Johan Karlsson$^{\dagger}$ \qquad Jian Li$^{\star}$\thanks{ 
This work was supported in part by NSF CCF-1218388, and Swedish Research Council. The views and conclusions contained herein are those of the authors and should not be interpreted as necessarily representing the official policies or endorsements, either expressed or implied, of the U.S. Government. The U.S. Government is authorized to reproduce and distribute reprints for Governmental purposes notwithstanding any copyright notation thereon.
Email:{\tt \{asadeghian, lemduck1\}@ufl.edu, johan.karlsson@math.kth.se, li@dsp.ufl.edu.}
}}
\address{$^{\star}$ University of Florida, Department of Electrical and Computer Engineering \\
    $^{\dagger}$ KTH Royal Institute of Technology, Department of Mathematics}
\begin{document}
\newtheorem{thm}{Theorem}
\newtheorem{cor}[thm]{Corollary}
\newtheorem{lemma}[thm]{Lemma}
\newtheorem{prop}[thm]{Proposition}
\newtheorem{problem}[thm]{Problem}
\newtheorem{remark}{Remark}
\newtheorem{defn}[thm]{Definition}
\newtheorem{ex}[thm]{Example}
\newcommand{\ignore}[1]{}
\newcommand{\mR}{{\mathbb R}}
\newcommand{\mN}{{\mathbb N}}
\newcommand{\mI}{{\mathbb I}}
\newcommand{\supp}{{\rm supp}}
\newcommand{\ext}{{\rm ext}}
\newcommand{\rR}{{\bold R}}
\newcommand{\rG}{{\bold G}}
\newcommand{\mZ}{{\mathbb Z}}
\newcommand{\mC}{{\mathbb C}}
\newcommand{\mT}{{\mathbb T}}
\newcommand{\fR}{{\mathfrak R}}
\newcommand{\fK}{{\mathfrak K}}
\newcommand{\fG}{{\mathfrak G}}
\newcommand{\fT}{{\mathfrak T}}
\newcommand{\fC}{{\mathfrak C}}
\newcommand{\fF}{{\mathfrak F}}
\newcommand{\cF}{{\mathcal F}}
\newcommand{\fN}{{\mathfrak N}}
\newcommand{\fM}{{\mathfrak M}}
\newcommand{\mD}{{\mathbb{D}}}
\newcommand{\cU}{{\mathcal{U}}}
\newcommand{\cL}{{\mathcal{L}}}
\newcommand{\cE}{{\mathcal{E}}}
\newcommand{\cI}{{\mathcal{I}}}
\newcommand{\cJ}{{\mathcal{J}}}
\newcommand{\cD}{{\mathcal{D}}}
\newcommand{\cR}{{\mathcal{R}}}
\newcommand{\cG}{{\mathcal{G}}}
\newcommand{\bJ}{{\mathbb{J}}}
\newcommand{\pC}{C_{\rm perio}((-\pi,\pi])}
\newcommand{\s}{{\rm s}}
\newcommand{\dtheta}{{\frac{d\theta}{2\pi}}}
\newcommand{\ar}{{\rm a}}
\newcommand{\intpi}{{\int_{-\pi}^\pi}}
\newcommand{\me}{{\rm _{ME}}}
\newcommand{\mr}{{\rm _{sm}}}
\def\spacingset#1{\def\baselinestretch{#1}\small\normalsize}
\newcommand{\bc}{{{\bf c}}}
\newcommand{\bd}{{{\bf d}}}
\newcommand{\Sg}{{{\bf S}_{\rm g}}}
\newcommand{\Sm}{{{\bf S}_{\rm m}}}
\newcommand{\Ng}{{N_{\rm g}}}
\newcommand{\Nm}{{N_{\rm m}}}
\newcommand{\Rg}{{{\bf R}_{\rm g}}}
\newcommand{\Rm}{{{\bf R}_{\rm m}}}
\newcommand{\yg}{{{\bf y}_{\rm g}}}
\newcommand{\ym}{{{\bf y}_{\rm m}}}
\newcommand{\yf}{{{\bf y}_{\rm f}}}
\newcommand{\ag}{{{\bf a}_{\rm g}}}
\newcommand{\am}{{{\bf a}_{\rm m}}}
\newcommand{\mO}{{\mathcal{O}}}
\newcommand{\mL}{{\mathcal{L}}}

\newcommand{\be}{{\bf e}}
\newcommand{\bx}{{\bf x}}
\newcommand{\by}{{\bf y}}
\newcommand{\bA}{{\bf A}}
\newcommand{\bX}{{\bf X}}
\newcommand{\bY}{{\bf Y}}

\newcommand{\br}{{\bf r}}
\newcommand{\bNoll}{{\bf 0}}
\newcommand{\but}{{\bf \tilde u}}
\newcommand{\bwt}{\tilde{{\bm \theta}}}
\newcommand{\bL}{{\bf L}}
\newcommand{\bT}{{\bf T}}
\newcommand{\bC}{{\bf C}}
\newcommand{\ba}{{\bf a}}
\newcommand{\bS}{{\bf S}}
\newcommand{\bQk}{{\bf Q}_k}
\newcommand{\bR}{{\bf R}}
\newcommand{\bD}{{\bf D}}
\newcommand{\bP}{{\bf P}}
\newcommand{\bp}{{\bf p}}
\newcommand{\bu}{{\bf u}}
\newcommand{\bw}{{\bm \theta}}
\newcommand{\balpha}{{\bm \alpha}}
\newcommand{\bGamma}{{\bm \Gamma}}
\newcommand{\INg}{{\bf I}_{\Ng}}
\newcommand{\INm}{{\bf I}_{\Nm}}
\newcommand{\IN}{{\bf I}_{N}}
\newcommand{\bI}{{\bf I}}

\newcommand{\diag}{{\textrm{diag}}}
\newcommand{\Span}{{\textrm{span}}}
\newcommand{\tr}{{\textrm{trace}}}
\newcommand{\rem}{{\bf{Remark:}}}
\newcommand{\real}{{\mathfrak{Re}}}
\newcommand{\imag}{{\mathfrak{Im}}}
\newcommand{\degr}{\deg}
\newcommand{\Pol}{{\textrm{Pol}}}
%
\maketitle
\begin{abstract}
The use of distances based on optimal transportation has recently shown promise for discrimination of power spectra. In particular, spectral estimation methods based on $\ell_1$ regularization as well as covariance based methods can be shown to be robust with respect to such distances. These transportation distances provide a geometric framework where geodesics corresponds to smooth transition of spectral mass, and have been useful for tracking. 

In this paper we investigate the use of these distances for automatic target recognition. 
We study the use of the Monge-Kantorovich distance compared to the standard $\ell_2$ distance
for classifying civilian vehicles based on SAR images. 
We use a version of the Monge-Kantorovich distance that applies also for the case where the spectra may have different total mass, and we formulate the optimization problem as a minimum flow problem that can be computed using efficient algorithms. 


\end{abstract}
\section{Introduction}
\label{sec:intro}

In our information society there is an ever increasing stream of images, and automatic processing is a key to analyze and utilize this information efficiently. It is therefore essential to quantify differences and similarities in images in a mathematically sound way. Estimation methods for radar and sonar imaging are often based on statistical quantities, and  it is therefore natural to demand that a ``small' change in the spectral content results in a small change in relevant statistical quantities. This is not the case for many standard metrics where a small shift in the frequency of a spectral line results in a significant change in, e.g., the $\ell_1$ or the $\ell_2$ norm of the spectral difference.  

In this paper we focus on the Monge-Kantorovich distance~\cite{Villani}, also known as the earth movers distance in the computer science community;  a distance which is rooted in optimal transport and which has shown promise for both tracking and classification \cite{DeGol2014Clustering, Haker2004Registration, Hoffman2004Multitarget, jiang2012distances,   muskulus2011wasserstein, Schmidt2014Automatic} and is a distance that is robust with respect to measurement error~\cite{GKT, KarlssonNing2014}. In particular, for data-direct high resolution spectral estimation methods such as sparse methods based on $\ell_1$-regularization~\cite{ChenDonoho1998,TanRobertsLi2011} 
the magnitude of the true solution can be robustly recovered if the error is quantified using the Monge-Kantorovich distance and the support of the true signal is sparse and with separated components \cite{KarlssonNing2014}. For these problems, the so-called dictionary is by necessity highly coherent and no useful bounds can be obtained in terms of the $\ell_p$ norms~\cite{Wohlberg2003}. The Monge-Kantorovic distance, does not just compare images point by point, but instead penalizes the total transport of mass. 
Also for covariance based methods, distances such as the Monge-Kantorovic distance have been shown to be robust with respect to measurement error and robustness bounds are computable  \cite{KarlssonGeorgiou2013}. 


In this paper we consider automatic target recognition (ATR) of vehicles, where the goal is to analyze a SAR image of a parking lot and determine if a given car in the parking lot is a sedan, a sports utility vehicle (SUV), or a van. We compare the recognition rate using the Monge-Kantorovic distance to the recognition rate obtained using the classical $\ell_2$ distance. Section~\ref{sec:Background} gives a background where the transportation distance is defined. In Section~\ref{sec:CompOfWassMet} we reformulate the optimization problem of computing the transportation distance as a minimum cost flow problem. In Section~\ref{sec:ProbSet} the automatic recognition problem is presented and we describe the classification procedure. Finally, the results are presented in  Section~\ref{sec:Res}, and  Section~\ref{sec:Sum} contains concluding remarks.

\section{Background}
\label{sec:Background}
The Monge-Kantorovich distance represents the minimal transportation cost of moving one ``mass'' distribution to another with specified cost of moving one unit amount of mass from one location to another \cite{Villani}.

Consider two $K$-dimensional element-wise non-negative vectors $f_0$ and $f_1$ that each represent a distribution of ``mass'' at the locations $x\in\Omega$. Let
$m(x_0,x_1)$ denote the amount of mass transported from location $x_0$ to location $x_1$, and we say that $M=(m(x_0,x_1))_{x_0,x_1\in \Omega}\in \mR^{K\times K}$ is a feasible transportation plan from $f_0$ to $f_1$ if the respective marginals are equal to $f_0$ and $f_1$, i.e., if $M$ is in the set

\begin{eqnarray*}
\Pi(f_0, f_1) & := & \Big\{M=(m(x_0,x_1))_{x_0,x_1} : m(x_0,x_1)\geq 0,\nonumber \\ 
              &&  \sum_{x_1\in \Omega}m(x_0,x_1)=f_0(x_0),\quad x_0\in \Omega \nonumber\\
              &&  \sum_{x_0\in \Omega}m(x_0,x_1)=f_1(x_1), \quad x_1\in \Omega\Big\}.\nonumber
\end{eqnarray*}

Let $c(x_0,x_1)$ represent the cost of transferring one unit of mass from location $x_0\in \Omega$ to location $x_1\in \Omega$, and define the matrix of transportation costs by $C:=[c(x_0,x_1)]_{x_0,x_1\in \Omega}\in \mR^{K\times K}$. 
Then the minimum cost of transporting mass with distribution $f_0$ to a distribution $f_1$ is
\begin{eqnarray}\label{eq:Tdef}
T_c(f_0, f_1)=\min_{M\in\Pi(f_0, f_1)} \sum_{x_0,x_1\in \Omega}m(x_0,x_1)c(x_0,x_1).
\end{eqnarray}
This is known as the Monge-Kantorovich distance \cite{Kantorovich1942}. 
Monge-Kantorovich distances are not metrics in general, but they
readily give rise to a class of the so-called Wasserstein metrics:
\[
W_{p,d}(f_0, f_1)= T_{c}(f_0, f_1)^{\min(1,\frac{1}{p})}
\]
where the cost function is of the form
$c({x_0},{x_1})=d({x_0},{x_1})^p$, and where $d$ is a metric on $\Omega$ and $p\in (0,\infty)$ \cite{Villani}. 

The Monge-Kantorovich theory deals with mass distributions of equal mass. However, they can be generalized to distances for distributions of possibly unequal masses as follows \cite{GKT}.
Given the two mass distributions $f_0$ and $f_1$, we postulate that these are perturbations of 
two other mass distributions $g_0, g_1\in \mR^K$, that have equal mass. Then, the cost of transporting ${f_0}$
and $f_1$ to one another can be thought of as the cost of transporting $g_0$ and $g_1$ to one another
plus the size of the respective perturbations:
\begin{equation}\label{eq:tildeT}
\tilde T_{c,\kappa}(f_0,f_1):=\hspace*{-5pt}\inf_{\|g_0\|_1=\| g_1\|_1}\hspace*{-5pt}T(g_0,g_1)+\kappa\sum_{j=0}^1 \|f_j-g_j\|_1.
\end{equation}

These distances have several interesting properties. They are weak$^*$ continuous hence may be used to localize spectral mass \cite{Villani, KarlssonGeorgiou2013}. They are contractive with respect to additive and normalized multiplicative noise, reflecting the fact that noise impedes the ability to discriminate.~\cite{GKT}. Furthermore, they have additional properties relating to deformations of spectra and smoothness with respect to translation. More specifically geodesics (e.g., the Wasserstein-2 metric) preserve ``lumpiness.'' A consequence of this is that when linking power spectra via geodesics of the metric, the corresponding peaks often seem to be  ``matched'' and the power between those transfer in a consistent manner. Such a property appears highly desirable in morphing for, e.g., tracking of frequencies in a slowly time-varying signal and integrating data from a variety of sources (see, e.g.,~\cite{JiangLuoGeorgiou2008,JiangTakyarGeorgiou2008, jiang2012distances}).
See also \cite{ning2014metrics} for a matrix valued extension.

\section{Computation of the Monge-Kantorovich distance} \label{sec:CompWass}
\label{sec:CompOfWassMet}
The computation of the Monge-Kantorovich distance is a linear optimization problem and can in principle be computed using any standard convex optimization software. 
We can write the Monge-Kantorovich distance~\eqref{eq:tildeT} as:

\begin{subequations}\label{eq:MinWass} 

\begin{eqnarray} \label{eq:MinWassA1} 
\hspace{-0.5cm}\tilde T_{c,\kappa}(f_0,f_1)&\!=\!& \min_{M,g_0,g_1} \!{\rm Tr}(M^TC)+ \kappa \sum_{i=0}^{1} \|f_i-g_i\|_1\label{eq:MinWassB}\\ 
\mbox{subject to} && 
{M \, {\mathbf{1}}_{K} = g_0 } \label{eq:MinWassC}\\
&&{M^T{\mathbf{1}}_{K} = g_1 } \label{eq:MinWassD}\\
&&{M \geq_{\rm e} 0 }
 \label{eq:MinWassA2}
\end{eqnarray}
\end{subequations}
where $M\in \mR^{K \times K}$ is a matrix that represents the transportation plan from $g_0$ to $g_1$, and $C=[c(x_i,x_j)]_{x_i,x_j\in \Omega}\in \mR^{K \times K}$ is the cost matrix that contains the costs of moving a unit of mass from one point to another. Here $\geq_{\rm e}$ denotes element-wise inequality and ${\mathbf{1}}_{K}$ is the $K\times 1$ vector of ones.

One challenge here is the computational burden of computing the distances for large $K$. However, it is well known that the optimal transport problem can be posed as a minimal cost flow problem (see, e.g., \cite{Klein1967}). We will here show that this approach may be modified to include the optimization problem \eqref{eq:tildeT}, hence allowing for the use of efficient specialized network algorithms for fast computations \cite{Orlin1996}.


\subsection{Monge-Kantorovich Distance as a Network Simplex Problem}
\label{sec:GraphRepr}
In this section we will describe how the Monge-Kantorovich distance~\eqref{eq:tildeT} can be formulated as a \textit{minimum cost flow problem}.
Finding the minimum-cost flow consists of determining the cheapest way to transport a given  supply to a given demand  through a graph, and such problems can be solved efficiently.

More specifically, a minimum-cost flow problem is formulated as follows. Let $G=(V,E)$ be a directed graph 
with a cost $\hat c(u,v)$ associated with each edge $(u,v) \in E$.
Then associate each node $v\in V$ with a number $d(v)\in \mR$ corresponding to the supply  of that node if $d(v)> 0$  and the demand of that node if $d(v)< 0$. 
The problem is then to find the flow, ${\varphi : E \rightarrow \mathbb{R}_{\geq 0} }$, that matches the supply to the demand with  minimal total cost: 
\begin{subequations}\label{eq:minflow}
\begin{eqnarray}
\hspace{-1cm}&& \underset{\varphi }{\text{minimize}}
\sum_{(u,v)\in E} \hat c(u,v)\varphi(u,v) \label{eq:minflowA}\\
\hspace{-1cm}&& \text{subject to}\nonumber \\
\hspace{-1cm}& & \sum \limits_{(v,u)\in E}  \varphi(v,u)-\hspace{-.3cm}\sum_{(u,v)\in E}  \varphi(u,v) = d(v)\text{ for } v \in V, \label{eq:minflowB}\\
\hspace{-1cm}& & \varphi(u,v)\ge 0 \mbox{ for } (u,v)\in E.\label{eq:minflowC}
\end{eqnarray}
\end{subequations}




Next, we will formulate~\eqref{eq:MinWass} as a minimum cost flow problem.
Let each of the two sets $\mathcal{F}_0=\{u_i:\, i=1,\ldots, K\}$ and $\mathcal{F}_1=\{v_i:\, i=1,\ldots, K\}$ correspond to the set of sample point of $x_i\in \Omega$, 
and let $d(v_i)=f_0(x_i)$ and $d(u_i)=-f_1(x_i)$, for $1 \leq i \leq K$, be the corresponding supply or demand.
Let $G_0=(V_0,E_0)$ be the complete bipartite di-graph with bipartition $\mathcal{F}_0$ and $\mathcal{F}_1$.   The cost of the edge connecting $u_i\in \mathcal{F}_0$ to $v_j\in \mathcal{F}_1$ is assigned as $\hat c(v_i,u_j )=c(x_i,x_j)$ in~\eqref{eq:MinWass}, i.e. the distance between $x_i$ and $x_j$.
The minimum cost flow problem \eqref{eq:minflow} corresponding to $G_0$ with costs $\hat c$ and demand/supply rates $d$ corresponds to the standard transportation problem \eqref{eq:Tdef}. 


In order to allow for mass perturbations \eqref{eq:tildeT} we will add an extra node.
To this end, let $G=(V,E)$ where $V=V_0\cup w$, and let $w$ be connected to every other node in $V_0$, i.e, $E=E_0\cup \{(w,v)\cup (v,w), v\in V_0\}$.
Further, let the cost of the edges be $\hat c(w,v)= \hat c(v,w)=\kappa$ for $v\in V_0$, and let the demand of $w$ be 
\[
d(w)=   \|f_1\|_1-\|f_0\|_1.
\]
By introducing this demand the total demand and supply add up to zeros also when $f_0$ and $f_1$ has different total mass.

One can easily see that the minimum cost flow of $G$ will equal to the transportation cost  $T_{\kappa,c}(f_0,f_1)$. In this setting, the functions $g_0$ and $g_1$ in \eqref{eq:MinWass} correspond to the supply and demand resulting from the flow in $G_0$, and  $\|g_i-f_i\|_1$ correspond to the flow between $w$ and $V_i$.

Solving the min-cost flow in a graph has been well studied previously starting with the early work of D. R. Fulkerson in 1961 \cite{fulkerson1961out}. A polynomial time network simplex algorithm for minimum cost flow problems has been given in \cite{Orlin1996}. Table~\ref{TimeTable} shows the time advantage of using this method compared to directly solving~\eqref{eq:MinWass} using a general purpose convex optimization tool like CVX.
\vspace{-3mm}
\begin{table}[h]
\centering
\caption{Time to compute $W_{\kappa,c}$ of two images using the two $~~~~~~~~~~~~~~~~$ algorithms (in seconds).}
\vspace{1.5mm}
\label{TimeTable}
\begin{tabular}{ c|c|c|c|c|c|c| }
\cline{2-7}
   {}    & \multicolumn{3}{c|}{29 $\times$ 24 pixels}  & \multicolumn{3}{c|}{58 $\times$ 48 pixels} \\
\cline{2-7}
   {}    & {$\kappa=1$} & {$\kappa=16$} &{$\kappa=32$} & {$\kappa=1$} & {$\kappa=16$} & {$\kappa=32$} \\ \hline
\multicolumn{1}{|c|}{CVX}   & {47.53}   & {47.79}  & {47.62}  & {698.6}  & {802.5} & {817.8}   \\ 
\hline
\multicolumn{1}{|c|}{CPLEX} & {0.018}  & {0.040} & {0.062} & {0.244} & {0.631} & {0.927}    \\ 
\hline
\end{tabular}
\end{table}
\vspace{-2mm}
\subsection{Role of $\kappa$}
The Monge-Kantorovich distance contains a free parameter $\kappa$ that specifies the penalty of adding and removing a unit of mass to the spectra.
In the reformulation of $\tilde T_{c,\kappa}$ as the min-cost flow problem, the flow of the optimal solution in any edge with cost greater than $2\kappa$ is going to be 0. To see this, assume that  $(\hat u,\hat v) \in E$ is an edge with $c(\hat u,\hat v)>2\kappa$ and $\varphi(\hat u,\hat v)>0$. Then the flow $\hat{\varphi}$ given~by
\begin{equation*} \label{eq:fhatdef}
\begin{aligned}
& \hat{\varphi}(u,v)=\varphi(u,v) \quad \mbox{ for all } u \in \mathcal{F}_0\backslash \{\hat u\}, v \in \mathcal{F}_1\backslash \{\hat v\} 
\\
&\hat{\varphi}(\hat u,w)=\varphi(\hat u,w)+\varphi(\hat u,\hat v) \\
&\hat{\varphi}(w,\hat v)=\varphi(w,\hat v)+\varphi(\hat u,\hat v) \\
&\hat{\varphi}(\hat u,\hat v)=0
\end{aligned}
\end{equation*}
is feasible and with lower cost:
\begin{eqnarray*}
&&\sum_{(u,v)\in E} \hat c(u,v)\varphi(u,v)-\sum_{(u,v)\in E} \hat c(u,v)\hat \varphi(u,v) \\
&&= \varphi(\hat u,\hat v)(\hat c(\hat u,\hat v)-2\kappa) >0.
\end{eqnarray*}
This contradicts that $\varphi$ is the minimum cost flow hence the support of $\varphi$ may be restricted to the edges of cost less or equal to $2\kappa$.

This observation significantly reduces the number of edges in the graph and hence reduces the computations required for calculating the Monge-Kantorovich distance. The computational time of the network simplex algorithm is $O(K^2 N^2  \log(K))$ where $K$ is the number of nodes in the graph and $N$ is the number of edges \cite{Orlin1996}. Therefore, if the use of $\kappa$ reduces the number of edges with a factor $p$, the computation time will be reduced by a factor of $p^2$.



\section{Automatic target recognition}
\label{sec:ProbSet}
We consider the problem of automatic target recognition of civilian vehicles and the goal is to analyze a SAR image of a parking lot and determine if a given car in the parking lot is a sedan, a SUV or a van. This recognition problem is solved by first identifying the vehicles and then using a classification method to determine which class the car belongs to. We compute the results both using the Monge-Kantorovich distance and the $\ell_2$ distance as distance for the classification method in order to compare recognition rates.

\subsection{The data set}

We use the Gotcha 2008 \cite{dungan2012wide} data set 
where SAR images are taken by an airborne radar from a circular flight pattern\footnote{We use the GOTCHA volumetric SAR data in this example from the U.S. Air force Sensor Data Management System. This data is publicly available by request.}. SAR imaging comes down to a 2D spectral estimation problem \cite{JakowatzWahlEichel1996} and gives an image of reflections for a given look angle. These are solved using sparse imaging methods \cite{TanRobertsLi2011} and then fused together using standard SAR imaging techniques.
This results in a data set containing images of 535 cars parked in a parking lot, 231 images of Sedans, 182 SUVs, 122 Vans. For the sake of an equal group size, 120 images are picked from each car type. 

\begin{figure}[h]
\centering
\includegraphics[width=7.5 cm]{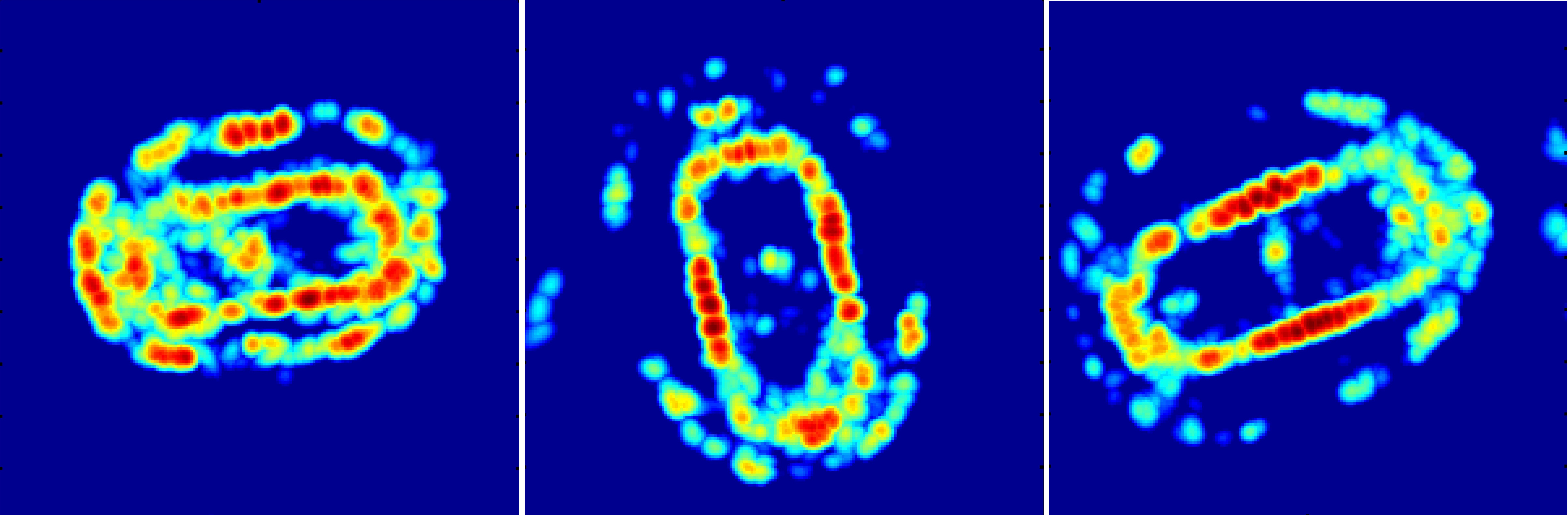}
\caption{SAR images of three different car classes. Sedan (left), SUV (center) and van (right).}
\label{fig:CarSAR} 
\end{figure}

As a preprocessing step, the cars in the image are rotated so that they are aligned and cropped such that each image contains only the car image. 
The pose estimation method is described in \cite{Lim2014Pose}. %
To speed up the computation of the Monge-Kantorovich distance, images are scaled down to $58 \times 48$ and $29 \times 24$ pixels. The rescaling uses a bicubic interpolation where each pixel is replaced by a weighted average of pixels in the nearest $4 \times 4$ neighborhood. This also allows us to study the robustness against image resolution.
\begin{figure}[h]
\centering
\includegraphics[width=7.5 cm]{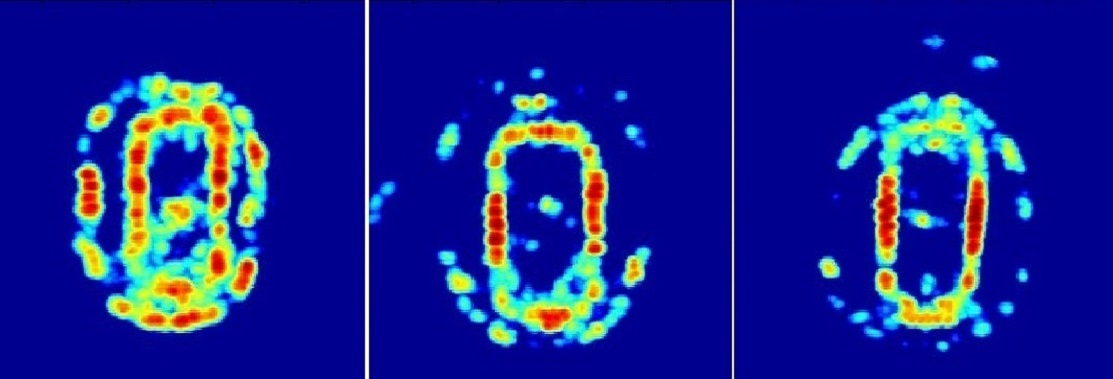}
\caption{SAR images of three different car classes after pose correction/scaling. Sedan (left), SUV (center) and van (right).}
\label{fig:CarSARPosed} 
\end{figure}

\vspace{-3mm}

\subsection{Methodology}
Next, the Monge-Kantorovich\footnote{The Euclidean distance $c(x,y)=\|x-y\|_2,$ where $ x,y\in \Omega$, is used as underlying distance.} and the $\ell_2$ distances are computed for all pairs of images. 
To solve the min-cost flow for the constructed graph in section~\ref{sec:GraphRepr} we used \small{TOMLAB CPLEX}  \normalsize\cite{TomlabCplex}. The solver is generally considered the state-of-the-art large scale mixed integer linear and quadratic programming solver.

For the classification, a training set is selected at random consisting of one third of the images from each group. The rest of the images are used as test data.
The test images are then classified using the nearest neighbour method, where each test image is associated with the class corresponding to the class of closest training image.
The error rate is then computed as the number of mislabeled cars divided by the total number~of~cars. This process is repeated $1000$ times and the average error rates are depicted in Fig.~\ref{fig:effectOfkappa}.

\section{Results} 
\label{sec:Res}
\begin{figure}
\centering
\includegraphics[width=7.5 cm]{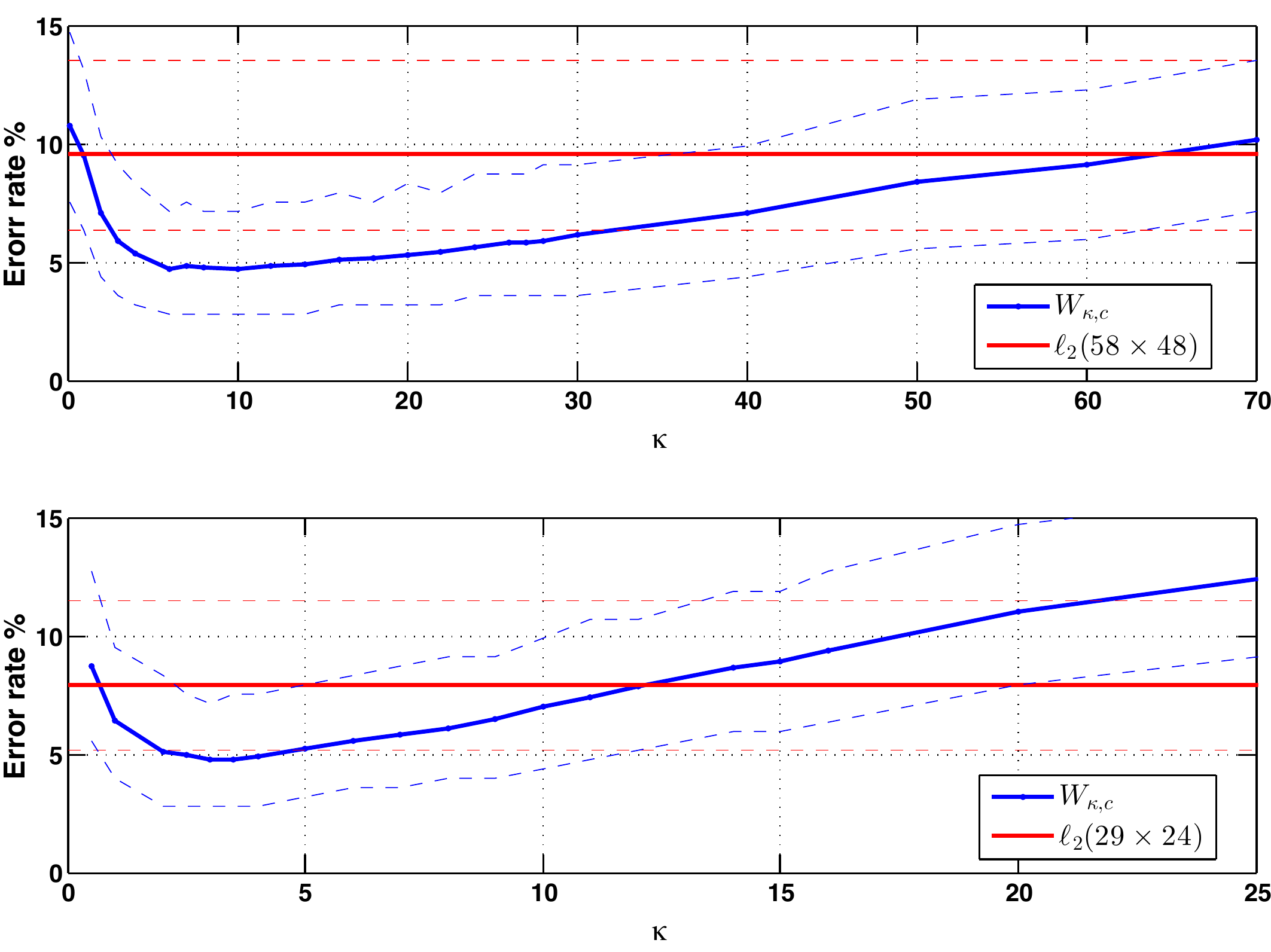}
\caption{The upper graph shows the error rate for $58 \times 48$ pixel images using the Wasserstein distance with different values of $\kappa$ as well as the error rate using $\ell_2$ as the distance. The lower graph shows the same results for $29 \times 24$ image size. Confidence intervals of level $90\%$ are shown in dashed lines.}
\label{fig:effectOfkappa} 
\end{figure}

From the error rates in Fig.~\ref{fig:effectOfkappa} it is seen that the recognition rates are considerably higher when using the Monge-Kantorovich distance compared to the $\ell_2$ distance provided that $\kappa$ is chosen appropriately.
Also as long as $\kappa$ is in a reasonable range, the recognition rate is not considerably sensitive to its value and hence it can be considered as a semi-parametric method.



From Fig.~\ref{fig:effectOfkappa} it can be seen that the optimal recovery rate for the Monge-Kantorovich based recognition is the same for the two image granularity levels  (for correctly selected $\kappa$). This suggests that this distance is relatively insensitive to rescaling/smoothing of the image. 
Also  note that when the image size is reduced, then the error rate for the $\ell_2$ norm is dropped by a considerable amount. The down-sampling method takes an average of the neighbouring pixels and use it as the new pixel value hence acts as smoothing. So when the $\ell_2$ norm is computed for the smaller image size, it is less sensitive to the pixel by pixel error and more sensitive to the the total spectral energy in a region. 


%
%
%
\section{Conclusions}
\label{sec:Sum}
In this paper we consider the optimal transport distance and its application for automatic target recognition. The results show that the error rate can be considerably lower when using the Monge-Kantorovich distance compared to the standard $\ell_2$ distance as underlying distance. 
We also present a fast way to compute the Monge-Kantorovich distance using the network simplex algorithm that applies also for spectra with different total mass.

%


\bibliographystyle{IEEEbib}
\bibliography{BibJka} 	

\begin{thebibliography}{10}

\bibitem{Villani}
C.~Villani,
\newblock {\em Topics in Optimal Transportation}, vol.~58,
\newblock Graduate studies in Mathematics, AMS, 2003.

\bibitem{DeGol2014Clustering}
J.~DeGol and M.~Nam,
\newblock ``A clustering approach for detecting moving objects captured by a
  moving aerial camera,''
\newblock in {\em Acoustics, Speech and Signal Processing (ICASSP), 2014 IEEE
  International Conference on}, May 2014, pp. 6538--6542.

\bibitem{Haker2004Registration}
S.~Haker, L.~Zhu, A.~Tannenbaum, and S.~Angenent,
\newblock ``Optimal mass transport for registration and warping,''
\newblock {\em International Journal of Computer Vision}, vol. 60, no. 3, pp.
  225--240, 2004.

\bibitem{Hoffman2004Multitarget}
J.R. Hoffman and R.P.S. Mahler,
\newblock ``Multitarget miss distance via optimal assignment,''
\newblock {\em Systems, Man and Cybernetics, Part A: Systems and Humans, IEEE
  Transactions on}, vol. 34, no. 3, pp. 327--336, May 2004.

\bibitem{jiang2012distances}
X.~Jiang, L.~Ning, and T.T. Georgiou,
\newblock ``Distances and riemannian metrics for multivariate spectral
  densities,''
\newblock {\em IEEE Transactions on Automatic Control}, vol. 57, no. 7, pp.
  1723--1735, 2012.

\bibitem{muskulus2011wasserstein}
M.~Muskulus and S.~Verduyn-Lunel,
\newblock ``Wasserstein distances in the analysis of time series and dynamical
  systems,''
\newblock {\em Physica D: Nonlinear Phenomena}, vol. 240, no. 1, pp. 45--58,
  2011.

\bibitem{Schmidt2014Automatic}
L.~Schmidt, C.~Hegde, P.~Indyk, J.~Kane, Ligang Lu, and D.~Hohl,
\newblock ``Automatic fault localization using the generalized earth mover's
  distance,''
\newblock in {\em Acoustics, Speech and Signal Processing (ICASSP), 2014 IEEE
  International Conference on}, May 2014, pp. 8134--8138.

\bibitem{GKT}
T.~Georgiou, J.~Karlsson, and M.S. Takyar,
\newblock ``Metrics for power spectra: An axiomatic approach,''
\newblock {\em IEEE Transactions on Signal Processing}, vol. 57, no. 3, pp.
  859--867, March 2009.

\bibitem{KarlssonNing2014}
J.~Karlsson and L.~Ning,
\newblock ``On robustness of $\ell_1$-regularization methods for spectral
  estimation,''
\newblock in {\em IEEE 53nd Annual Conference on Decision and Control (CDC)},
  Dec 2014.

\bibitem{ChenDonoho1998}
S.~Chen and D.~Donoho,
\newblock ``Application of basis pursuit in spectrum estimation,''
\newblock in {\em IEEE International Conference on Acoustics, Speech, and
  Signal Processing}, 1998, vol.~3, pp. 1865--1868.

\bibitem{TanRobertsLi2011}
X.~Tan, W.~Roberts, J.~Li, and P.~Stoica,
\newblock ``Sparse learning via iterative minimization with application to
  {MIMO} radar imaging,''
\newblock {\em IEEE Transactions on Signal Processing}, vol. 59, no. 3, pp.
  1088--1101, March 2011.

\bibitem{Wohlberg2003}
B.~Wohlberg,
\newblock ``Noise sensitivity of sparse signal representations: reconstruction
  error bounds for the inverse problem,''
\newblock {\em IEEE Transactions on Signal Processing}, vol. 51, no. 12, pp.
  3053--3060, Dec 2003.

\bibitem{KarlssonGeorgiou2013}
J.~Karlsson and T.~Georgiou,
\newblock ``Uncertainty bounds for spectral estimation,''
\newblock {\em IEEE Transactions on Automatic Control}, vol. 58, no. 7, pp.
  1659--1673, July 2013.

\bibitem{Kantorovich1942}
L.V. Kantorovich,
\newblock ``On the transfer of masses,''
\newblock in {\em Dokl. Akad. Nauk. SSSR}, 1942, vol.~37, pp. 227--229.

\bibitem{JiangLuoGeorgiou2008}
X.~Jiang, Z.~Luo, and T.T. Georgiou,
\newblock ``Power spectral geodesics and tracking,''
\newblock in {\em 47th IEEE Conference on Decision and Control}, December 2008,
  pp. 1315--1319.

\bibitem{JiangTakyarGeorgiou2008}
X.~Jiang, S.~Takyar, and T.T. Georgiou,
\newblock {\em Metrics and morphing of power spectra}, vol. 371 of {\em Lecture
  Notes in Control and Information Sciences}, pp. 125--135,
\newblock Springer Verlag, 2008,
\newblock (V. Blondel and S. Boyd and H. Kimura, eds.).

\bibitem{ning2014metrics}
L.~Ning and T.T. Georgiou,
\newblock ``Metrics for matrix-valued measures via test functions,''
\newblock {\em arXiv preprint arXiv:1409.4097}, 2014.

\bibitem{Klein1967}
M.~Klein,
\newblock ``A primal method for minimal cost flows with applications to the
  assignment and transportation problems,''
\newblock {\em Management Science}, vol. 14, no. 3, pp. 205--220, 1967.

\bibitem{Orlin1996}
J.~B. Orlin,
\newblock ``A polynomial time primal network simplex algorithm for minimum cost
  flows,''
\newblock pp. 474--481, 1996.

\bibitem{fulkerson1961out}
D.R. Fulkerson,
\newblock ``An out-of-kilter method for minimal-cost flow problems,''
\newblock {\em Journal of the Society for Industrial \& Applied Mathematics},
  vol. 9, no. 1, pp. 18--27, 1961.

\bibitem{dungan2012wide}
K.~E. Dungan, J.~N. Ash, J.~W. Nehrbass, J.~T. Parker, L.~A. Gorham, and S.~M
  Scarborough,
\newblock ``Wide angle {SAR} data for target discrimination research,''
\newblock in {\em SPIE Defense, Security, and Sensing}. International Society
  for Optics and Photonics, 2012, pp. 83940M--83940M.

\bibitem{JakowatzWahlEichel1996}
C.~V. Jakowatz, D.~E. Wahl, P.~H. Eichel, D.~C. Ghiglia, and P.~A. Thompson,
\newblock {\em Spotlight-mode Synthetic Aperture Radar: A Signal Processing
  Approach},
\newblock Springer Science and Business Media, Inc., 1996.

\bibitem{Lim2014Pose}
D.~Lim, L.~Xu, Y.~Sun, and J.~Li,
\newblock ``Wide-angle high resolution {SAR} imaging and robust automatic
  target recognition of civilian vehicles,''
\newblock in {\em International Journal of Remote Sensing Applications}. To
  appear.

\bibitem{TomlabCplex}
{IBM }Corp,
\newblock ``{TOMLAB CPLEX},'' http://tomopt.com/to ... mlab/products/cplex/,
  Version 7.9.

\end{thebibliography}

\end{document}